%% file: formatting-instructions-latex-2020.tex
\def\year{2020}\relax
\documentclass[letterpaper]{article} 
\usepackage{aaai20}  
\usepackage{times}  
\usepackage{helvet} 
\usepackage{courier}  
\usepackage[hyphens]{url}  
\usepackage{graphicx} 
\usepackage{graphicx}  
\usepackage{bm}  
\usepackage{amsmath,amssymb,amstext, mathrsfs,multirow}
\usepackage{wasysym}
\usepackage{tikzsymbols}
\newcommand{\E}{\operatorname{\mathbb{E}}}
\newcommand{\citet}[1]{\citeauthor{#1}~\shortcite{#1}} \newcommand{\citep}{\cite}

\title{Generating Emotionally Aligned Responses in Neural Dialogue Systems\\ using Affect Control Theory}
\author{Nabiha Asghar\textsuperscript{\rm *}\textsuperscript{\rm 1}\textsuperscript{\rm 2} Ivan Kobyzev\textsuperscript{\rm *}\textsuperscript{\rm 1} Jesse Hoey\textsuperscript{\rm 1}\textsuperscript{\rm 2} Pascal Poupart\textsuperscript{\rm 1}\textsuperscript{\rm 2} Muhammad Bilal Sheikh\\ 
\textsuperscript{\rm *}Equal Contribution\\
\textsuperscript{\rm 1}Cheriton School of Computer Science, Univerity of Waterloo, Canada\\ 
\textsuperscript{\rm 2}Vector Institute for AI, Canada\\
\texttt{\{nasghar,ikobyzev,jhoey,ppoupart\}@uwaterloo.ca}\\
\texttt{mbsheikh298@gmail.com}
}
 
\begin{document}

\maketitle

\input{bayesactdefs}

\begin{abstract}
State-of-the-art neural dialogue systems excel at syntactic and semantic modelling of language, but often have a hard time establishing emotional alignment with the human interactant during a conversation. In this work, we bring Affect Control Theory (ACT), a socio-mathematical model of emotions for human-human interactions, to the neural dialogue generation setting. ACT makes predictions about how humans respond to emotional stimuli in social situations. Due to this property, ACT and its derivative probabilistic models have been successfully deployed in several applications of Human-Computer Interaction, including empathetic tutoring systems, assistive healthcare devices and two-person social dilemma games. We investigate how ACT can be used to develop \textit{affect-aware} neural conversational agents, which produce emotionally aligned responses to prompts and take into consideration the affective identities of the interactants. 

\end{abstract}

\section{Introduction}
\label{bact_gen_intro}

In the rapidly evolving field of text-based human-computer interaction (HCI), there is an increasing focus on developing dialogue systems\footnote{Also known as conversational agents, or virtual assistants, or chatbots.} that are emotion/affect aware. Affectively cognizant conversational agents have been shown to provide companionship to humans \cite{prendinger2005empathic,catania2019emoty}, help improve emotional wellbeing~\cite{ghandeharioun2018emma}, give medical assistance in a more humane way \cite{malhotra2015exploratory}, help students learn efficiently \cite{kort2001affective}, and assist mental healthcare provision to alleviate bullying~\cite{gordon2019primer}, suicide and depression \cite{jaques2017multimodal,taylor2017personalized}. The importance of the agent's affect awareness is obvious in open-domain dialogue (e.g. for entertainment or companionship). In addition, task-oriented settings like customer service can also benefit from virtual agents that are responsive towards implicit or explicit emotional cues from the user, such as expressing dissatisfaction about a product or negotiating price.
 
Recent breakthroughs in natural language processing (NLP) have significantly advanced the state-of-the-art in emotion-aware text-based dialogue generation. Several neural-network based affective dialogue models have been explored in the literature~\cite{shen2017conditional,zhang2018learning}. 
However, most of these systems suffer from one or more of the following challenges. First, they model emotion as a set of discrete categories. This is a prohibitive assumption, because humans often experience emotions as a continuum (i.e., a mixture of several feelings of varying intensity) rather than a single emotion of  fixed  intensity. 
Second, these studies do not take into account the affective identity of the user during the interaction. For instance, a conversation happening between two friends would typically be very different from the one between two enemies, but this is not accounted for by most modern systems.

To address these limitations, we propose to augment neural dialogue models with Affect Control Theory~\cite[ACT]{Heise2007}, a socio-mathematical model of affect. ACT  models the affective/emotional aspects of social interactions between two humans. Given the affective \textit{identity} of each of the two interactants, ACT prescribes affective actions for them that are mutually aligned towards minimizing conflict. Since ACT is primarily a theory of interactions, it lends itself naturally to the dialogue setting. We augment text-based dialogue agents with the ability to reason about affect using ACT. In doing so, we enable them to perceive human emotions (conveyed through the text) and produce emotionally appropriate textual responses based in an affective context of identities.

ACT and neural dialogue models have fundamentally different representation spaces, thus integrating them is not straightforward. ACT operates in a 3-dimensional continuous affective space, where the basis vectors are Evaluation (E), Potency (P) and Activity (A). On the other hand, actions in a dialogue system are typically sentences that convey some affect as well as one or more propositions.
For instance, the sentences ``\textit{Could you please make me some tea}" and \textit{``Go make me some tea"} convey the same propositional action (asking for tea) but their affect is vastly different. The former can be seen as a request or appeal, 
whereas the latter is more of a command. 

\begin{figure*}
\centering
\includegraphics[width=0.75\linewidth]{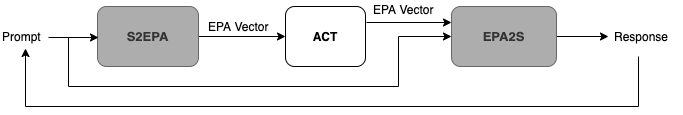}
\caption{Pipeline to integrate Affect Control Theory (ACT) into a dialogue system. The two components \texttt{S2EPA} and \texttt{EPA2S} are depicted as blackboxes, and are described later in the paper. }
\label{bactgen_pipeline}
\end{figure*}

To explore ACT's viability for text-based dialogue generation, and to deal with the representation-space discrepancy, we propose a neural encoder-decoder dialogue pipeline shown in Figure~\ref{bactgen_pipeline}. For a given input sentence/prompt, a sentence-to-EPA (\texttt{S2EPA}) function maps the input to an EPA vector, such that the vector appropriately conveys the affect of the input sentence. ACT is queried with this vector, and produces the response EPA vector. Then an EPA-to-sentence (\texttt{EPA2S}) function maps this response EPA, as well as the input prompt, to an output response, generated word by word, that is semantically relevant to the prompt and conveys the affect of the response EPA.
To the best of our knowledge, we are the first to bring ACT to the domain of dialogue generation. 

\section{Related Work}
\label{bact_gen_relatedwork}

Most of the early affective dialogue systems were retrieval-based or slot-based, and used hand-crafted speech and text-based features~\cite{callejas2011predicting,hasegawa2013predicting,pittermann2010emotion}. More recently, with the advent of sophisticated and highly flexible neural network models~\cite{serban2017hierarchical,shang2015neural,sutskever2014sequence,vinyals2015neural}, the focus has shifted to building data-driven end-to-end dialogue models. Retrieval-based systems are still popular because they are more controllable, require less training data and are more efficient~\cite{gordon2019primer,huang2017moodswipe,zhou2018design}. However, generative models dominate this space because they generalize well~\cite{rashkin2018know,vadehra2018creating}. This work falls in the latter category.

A large part of the affective dialogue literature treats \textit{emotion} as a set of discrete categories, where each category corresponds to a type of biological response. For instance, some studies focus on producing \textit{sentiment}-appropriate responses, where sentiment refers to \textit{positive, negative} or \textit{neutral} emotion~\cite{kong2019adversarial,shi2018sentiment}. Other works use a larger set of discrete emotions~\cite{dryjanski2018affective,ghosh2017affectlm,zhang2018learning,zhou2017ecm}, based on the different psychological theories of emotion~\cite{ekman1992argument,plutchik1980general}. A recent research trend, propelled by social media growth, is to categorize emotions using the \textit{emoji}\footnote{An emoji is a symbol of emotional expression, such as a smiling/frowning face, a flower, etc.} spectrum. This enables model training using massive weakly labelled datasets, e.g., from Twitter~\cite{park2018finding,xie2016neural,zhou2018mojitalk}. For instance, ~\citet{fung2018empathetic} and  \citet{park2018finding} train emotion embeddings on tweets with hashtags and emojis as labels. These embeddings can be used downstream in other NLP tasks, such as dialogue systems. 

In recent years, several Seq2Seq-based affective conversational models have been proposed. 
Emotional Chatting Machine~\cite[ECM]{zhou2017ecm} takes as input a prompt and the desired emotion category of the response, and produces a response. ECM operates on 8 discrete emotion categories, has an internal memory that encodes how much an emotion has already been expressed, and an external memory that decides whether to choose an emotional or generic (non-emotional) word at a given step during decoding.
\citet{dryjanski2018affective} inject predefined sentiment to a neutral utterance by inferring the phrases and their insertion points. \citet{lubis2018eliciting} jointly train a Seq2Seq model and an emotion encoder. The emotion encoder maintains the emotional context during a conversation, and is trained using the SEMAINE dataset (2000 samples)~\cite{mckeown2012semaine} where utterances are labeled on the valence and arousal axes. \citet{asghar2018affective} use a continuous, three dimensional representation of emotions, which is used to augment pretrained word embeddings, training objectives, and beam search inference.  \citet{vadehra2018creating} train Seq2Seq with an adversarial objective to remove affect from the learned representation of the input utterance, and generate the response based on this representation and the target affect label (one of seven discrete emotion categories).
 \citet{rashkin2018know} have released \textit{EmpatheticDialogues}, 
a dataset of 25000 conversations grounded in emotional situations to facilitate training and evaluation of dialogue systems. They show that finetuning existing dialogue models on this dataset boosts their affective quality significantly.


Conditional Variational Autoencoders~\cite[CVAEs]{sohn2015learning} have become another popular choice for neural dialogue models. CVAEs have recently been used for affect-controlled dialogue generation, where the model is conditioned on positive-negative-neutral sentiment tags~\cite{shen2017conditional} or more fine-grained emotion categories ~\cite{zhang2018learning}.  \citet{kong2019adversarial} use an adversarial approach for sentiment control which can be applied to CVAEs too.

In this work, we follow \cite{asghar2018affective} and use a continuous, three dimensional representation of emotions. The three dimensions are Evaluation, Potency and Activity, and have been validated by several pioneering research studies in psychology~\cite{Osgood1975,Russell1977,Heise1979,Russell2003}. Intuitively, a continuous and multi-dimensional representation of emotions makes sense; as humans we experience emotions as a mixture of several feelings of varying intensity, rather than a single emotion of fixed intensity. Moreover, continuous emotion vectors fit well with dialogue models that are trained end-to-end. Using this 3D representation of emotions, we propose 1) a Seq2Seq based model inspired from ~\cite{asghar2018affective}, and 2) a CVAE based model; it is inspired from~\cite{shen2017conditional} and~\cite{zhang2018learning}, but leverages Affect Control Theory as an external model of affect for conditional response generation.

\section{Affect Control Theory}
Affect Control Theory (ACT) arises from work on the psychology and sociology of human social interaction~\cite{Heise2007}. ACT proposes that social perceptions, behaviours, and emotions are guided by a psychological need to minimize the differences between culturally shared fundamental  sentiments about social situations and the transient impressions resulting from the interactions between elements within those situations. Fundamental sentiments, $\fub$, are representations of social objects, such as interactants' identities and behaviours, as vectors in a 3D affective space, hypothesised to be a universal organising principle of human socio-emotional experience~\cite{Osgood1975}. The basis vectors of affective space are called Evaluation/valence, Potency/control, and Activity/arousal (EPA). EPA profiles of concepts can be measured with the {\em semantic differential}, a survey technique where respondents rate affective meanings of concepts on numerical scales with opposing adjectives at each end (e.g., good, nice vs. bad, awful for E, weak, little vs. strong, big for P, and calm, passive vs. exciting, active for A). Affect control theorists have compiled lexicons of a few thousand  words along with average EPA ratings obtained from survey participants who are knowledgeable about their culture~\cite{Heise2010}. For example, most English speakers agree that professors are about as nice as students (E), more powerful (P) and less active (A). The corresponding EPAs are $[1.7,1.8,0.5]$ for professor and $[1.8,0.7,1.2]$ for student\footnote{\label{foot:actlex} Unless otherwise noted, all EPA labels and values in the paper are taken from the Indiana 2002-2004 ACT lexicon~\cite{Heise2010}. Values range by historical convention from $-4.3$ to $+4.3$.}.  In Japan, professor has the same P ($1.8$) but students are seen as less powerful ($\mi 0.21$).\footnote{Taken from the Japan 1989-2002 dataset~\cite{Japan2002Data}}

Social events cause transient impressions, $\trb$ (also three dimensional in EPA space) of identities and behaviours that may deviate from their corresponding fundamental sentiments, $\fub$. ACT models this formation of impressions from events presented as  triples actor-behaviour-object. Consider, for example, a professor (actor) who yells (behaviour) at a student (object). Most would agree that this professor appears considerably less nice (E), a bit less potent (P), and certainly more aroused (A) than the cultural average of a professor. Such transient shifts in affective meaning caused by specific events are described with models of the form $\trb'=\ACTM\Gop(\fub,\trb)$, where $\ACTM$ is a matrix of statistically estimated prediction coefficients from empirical impression-formation studies and $\Gop$ is a vector of polynomial features in $\fub$ and $\trb$.
In ACT, the weighted Euclidean distance between fundamental sentiments and transient impressions is called {\em deflection} $d=\| \fub-\trb'\|_w^2$, and is hypothesised to correspond to an aversive state of mind that humans seek to avoid.  This {\em affect control principle} allows ACT to compute {\em prescriptive} actions for humans: those that minimize the deflection.  Emotions in ACT are computed as a function of the difference between fundamentals and transients~\cite{Heise2007}, and are thought to be communicative signals of vector deflection that help maintain alignment between cooperative agents. 

For two given identities of the actors (two EPA vectors) and an initial EPA action by one actor, ACT predicts the optimal response for the second actor through prediction equations~\cite{Heise2007,asghar2015intelligent}. For example, let the two identities be \textit{tutor} \epa{1.5}{1.4}{-0.2} and \textit{student} \epa{1.5}{0.3}{0.7}.
Let the initial action by \textit{tutor} be \textit{compromise with}\footnote{We can choose this initial action manually.}. Then, at time step 1, we query ACT with the event (\textit{tutor}, \textit{compromise with}, \textit{student}). The output of ACT's prediction equations is the optimal action\footnote{The process of computing the optimal action in ACT is described in detail in Chapters 11 and 12 of \cite{Heise2007}. Briefly, this involves computing the event likelihood, which is a quadratic function of the unknown weights. The optimal behaviour can be obtained by setting the partial derivatives of the likelihood to zero and solving for the behaviour terms.} \textit{student} should take. In this case, it is a 3D vector corresponding to the action \textit{confer with}. At the next time step, we can query ACT with the event (\textit{student, confer with, tutor}) to predict the optimal action to be taken by \textit{tutor} (in this case \textit{counsel} \epa{2.0}{1.5}{-0.5}).
In this way, ACT can be queried sequentially to carry out long interactions. 

ACT's predictions can be explored through computer simulations, via a freely available software called INTERACT\footnote{The ACT software, called INTERACT, is publicly available at \url{http://www.indiana.edu/~socpsy/ACT/interact.htm}.}.

\section{Proposed Model}

An overview of the proposed ACT conversational model is shown in Figure~\ref{bactgen_pipeline}. ACT is instantiated with two affective identities, one each for the human participant and the artificial agent. Given an input prompt (a sentence), a sentence-to-EPA (\texttt{S2EPA}) function maps the prompt to an EPA vector. This EPA vector acts as the affective action for one of the interactants in ACT, and ACT produces the EPA vector of the affective action taken by the other participant. This target EPA vector, together with the prompt, is used to generate a response sentence using an EPA-to-sentence (\texttt{EPA2S}) function. This response can be treated as the next prompt in the conversation, and the process continues.

\begin{figure*}
\centering
\includegraphics[width=0.7\linewidth]{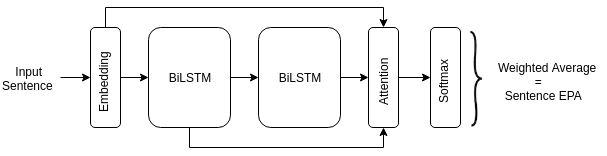}
\caption{\texttt{S2EPA}: A pretrained BiLSTM network with attention \cite{felbo2017using}, tweaked to produce EPA vectors instead of emojis.}
\label{deepmoji_arch}
\end{figure*}

We propose the following strategies to build the \texttt{S2EPA} and \texttt{EPA2S} functions.
\begin{enumerate}
    \item \textbf{\texttt{S2EPA}}: To map sentences to the 3-dimensional EPA space, we modify the output of a pretrained and publicly available BiLSTM network called DeepMoji \cite{felbo2017using}, which produces a probability distribution over a set of 64 \textit{emojis} given an input sentence. We achieve this by manually labeling the 64 emojis with EPA vectors, and taking a weighted average (using the softmax probabilities) of these vectors. We are making the assumption here that the same EPA would be generated by translating a sentence into a semantic behaviour label (e.g. translating "Go make me some tea" into the label "command" and thereby to a negative and powerful EPA), as will be generated by translating a sentence into an emoji (e.g. "Go make me some tea" with an angry emoji, and thereby to the same negative and powerful EPA). 
    This is shown in Figure~\ref{deepmoji_arch}. 
    \item \textbf{\texttt{EPA2S}}: To generate a sentence given an input prompt and a target EPA vector, we explore two models, traditional Seq2Seq~\cite{sutskever2014sequence} with attention and a conditional variational autoencoder (CVAE)~\cite{sohn2015learning}. In Seq2Seq, the target EPA and input response are passed through the encoder together to produce a fixed-length context vector. This context is passed through the decoder to generate a response. On the other hand, the CVAE model encodes the input into a Gaussian latent space. A sample from this latent space is then propagated through a decoder to generate an appropriate response.

\end{enumerate}

\noindent
We now describe each of \texttt{S2EPA} and \texttt{EPA2S} in detail. 

\subsection{Sentence to EPA (\texttt{S2EPA})}
The goal of \texttt{S2EPA} function is to generate an EPA representation of a given sentence. If we had access to a large amount of sentences labeled with EPAs, we could simply train a recurrent neural network to approximate the sentence-to-EPA mapping. However, building such a dataset is time-consuming and expensive.
The other option is to use the word-level EPA values, but then semantic understanding of the sentence is required. That is, a sentence needs to be parsed into actor-behaviour-object triples~\cite{alhothali2017semi}. To get around this issue, we use a pre-trained publicly available sentence-to-emoji model and tweak its output to suit our needs.

Concretely, we use DeepMoji, a pretrained BiLSTM network with attention~\cite{felbo2017using}\footnote{The pretrained DeepMoji model is publicly available at \url{https://github.com/bfelbo/DeepMoji}.}. This model is trained on a dataset of 1.2 billion tweets labeled with emojis. Given an input sentence, the model produces a probability distribution over 64 emojis. We use this model to our advantage as follows. We ask two human annotators to label these 64 emojis with EPA vectors. We average these annotations to assign a single EPA vector to each emoji. Then, given an input query, we take the weighted average of the 64 EPA vectors, where the weights are produced by the softmax layer. This gives us the desired sentence to EPA mapping. 
The architecture of \texttt{S2EPA} is shown in Figure~\ref{deepmoji_arch}.

\subsection{EPA to Sentence (\texttt{EPA2S})}
The goal of \texttt{EPA2S} function is to generate a response sentence, given the input prompt and a target EPA vector, such that the response conveys the same affect as the target EPA. To build \texttt{EPA2S}, we explore two methods, Seq2Seq and CVAE.

\subsubsection{\texttt{EPA2S-Seq2Seq}}
One straightforward model for \texttt{EPA2S} is Seq2Seq with attention, where the input sentence is concatenated with the target EPA and passed into the encoder. This produces a fixed-length context vector. Given this context vector, the decoder sequentially produces the response while attending to the encoder's hidden states.

\subsubsection{\texttt{EPA2S-CVAE}}
\begin{figure*}
\centering
\includegraphics[width=0.7\linewidth]{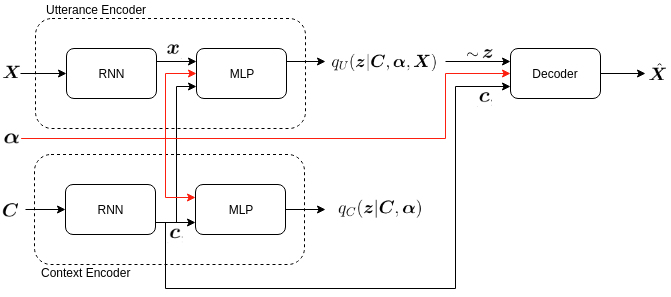}
\caption{CVAE training architecture.}
\label{cvae_model}
\end{figure*}

CVAE is another viable model for \texttt{EPA2S}. For the training set one has a collection of triples of the form  $(\bm C, \bm \alpha, \bm X)$, where  $\bm C$ and $\bm X$ are sequences of tokens denoting the prompt and the response respectively, and $\bm \alpha$ is an EPA vector of the response $\bm X$. The CVAE consists of a \textit{context encoder, utterance encoder} and a \textit{decoder}. The context encoder uses an RNN to map $\bm C$ to a fixed-length vector $\bm c$, and then passes $(\bm c, \bm \alpha)$ to an MLP, which outputs the parameters of the probability distribution $q_C(\bm z|\bm C, \bm \alpha) \sim \mathcal{N}(\bm \mu, \bm \lambda^2 \bm I)$; this distribution is called the prior. Similarly, the utterance encoder uses an RNN to map $\bm X$ to a fixed-length vector $\bm x$, and then passes $(\bm c, \bm \alpha, \bm x)$ to an MLP that outputs the parameters of the probability distribution $q_U(\bm z|\bm C, \bm \alpha, \bm X) \sim \mathcal{N}(\hat{\bm \mu}, \hat{\bm \lambda}^2 \bm I)$. This is the posterior. A latent vector $\bm z$ is then sampled from $q_U$. The decoder RNN parameterizes the distribution $q_D(\bm X | \bm z, \bm C, \bm \alpha)$; it takes $(\bm z, \bm c, \bm \alpha)$ as input and produces a distribution over the response sequences. The CVAE objective is to maximize the reconstruction probability of $\bm X$, and minimize the KL divergence between the prior $q_C$ and the posterior $q_U$. This is given by
\begin{align}
    L_{\texttt{CVAE}}\big(\bm \theta_C, \bm \theta_U, & \bm \theta_D; \bm C, \bm X, \bm \alpha \big) = \nonumber\\
    & \text{KL}\big(q_U(\bm z|\bm C, \bm \alpha, \bm X) \big\|q_C(\bm z|\bm C, \bm \alpha)\big)\nonumber \\
    & - \E_{q_U} \big[\log q_D(\bm X| \bm z, \bm C, \bm \alpha) \big]
    \label{cvae_loss}
\end{align}
where $\bm \theta_C, \bm \theta_U$ and $\bm \theta_D$ denote the parameters of the context encoder, the utterance encoder, and the decoder respectively. This training process is depicted in Figure~\ref{cvae_model}.

\begin{figure*}
\centering
\includegraphics[width=0.7\linewidth]{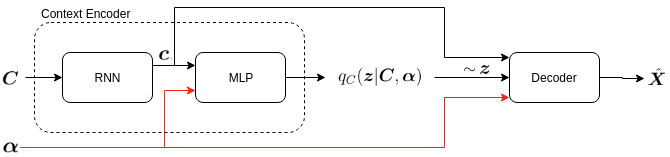}
\caption{CVAE at inference time: this is the  \texttt{EPA2S} function.}
\label{epa2s_fig}
\end{figure*}

For inference, the goal is to generate a response given an input sentence $\bm C$ and a target EPA $\bm \alpha$. $(\bm C, \bm \alpha)$ are passed through the context encoder, and a latent variable $\bm z$ is sampled from $q_C$. Then $(\bm z, \bm c, \bm \alpha)$ are passed to the decoder to generate a response. This process is depicted in Figure~\ref{epa2s_fig}.

\section{Experiments}
\subsection{Training, Data and Setup}
The Seq2Seq model with attention contains a single layer BiGRU network as the encoder, and a single layer GRU network as the decoder, each layer containing 300 cells. For the CVAE model, each encoder contains 1) a single-layer BiGRU, each direction containing 300 GRU cells, and 2) a two-layer MLP. The CVAE decoder is a single-layer GRU network of 300 cells. The variables $\bm z$ and $\bm \alpha$ are 300-dimensional and 3-dimensional respectively. 
 
 For both models, the vocabulary size is fixed to 24000 and the embedding layer is initialized with 300-dimensional GloVe embeddings~\cite{pennington2014glove}. The models are implemented in PyTorch 0.4 and optimized with Adam~\cite{adam} with an initial learning rate of $10^{-4}$ and other default parameters.
For training, we use the Cornell Movie Corpus~\cite{cornellcorpus}, which contains $\sim$220k prompt-response pairs from movie conversations. We split the data into 200k, 10k and 10k samples for training, validation and testing. For a given pair of ACT identities $id_1$ and $id_2$, we construct the training data as follows\footnote{See the sections on \textbf{Experiment \#3} and \textbf{Discussion} for more details on how to choose these identities, and why this choice matters.}. For each conversation in the Cornell corpus, we assume that the two identities say the utterances alternately. Then, for each training sample $(\bm C, \bm X)$ in the corpus, we query ACT with the event ($id_1$, \texttt{S2EPA}($\bm C$), $id_2$) or ($id_2$, \texttt{S2EPA}($\bm C$), $id_1$) depending on who uttered $\bm C$. This gives us the optimal response EPA. We include this target EPA vector $\bm \alpha$ to the training sample to obtain the triple $(\bm C, \bm \alpha, \bm X)$.

For the CVAE model, we follow~\cite{kingma2013auto}; we compute the  reconstruction loss with a single sample from $q_C$, and compute the KL divergence in closed form. Furthermore, to prevent the \textit{degenerate} case where the KL divergence is equal to zero, we use KL annealing, following ~\cite{bowman2016generating}. Degeneracy occurs when the network sets the posterior $q_U$ to be equal to the prior $q_C$, implying that the network ignores the latent variable. This is sometimes referred to as the \textit{vanishing latent variable problem}. KL annealing circumvents this issue by adding a weight to the KL term during training. In the beginning, this weight is zero, so the network encodes useful information in $\bm z$ without worrying about staying close to the prior. As training progresses, the weight is slowly increased till it reaches one. 


\subsection{Evaluation}
Existing automated dialogue evaluation metrics are not suitable for assessing the quality of an open-domain and affective conversational model~\cite{asghar2018affective,liu2016hownotto}. It is also unclear how to evaluate affective aspects by automated metrics. Therefore, we recruit human judges to evaluate the proposed models, following previous
studies~\cite{mou2016sequence,shang2015neural}.\\

\noindent
We carry out the following three experiments.

\subsubsection{Experiment \# 1}
First, we assess the quality of EPA vectors produced by the \texttt{S2EPA} model.
Some example sentences from the Cornell test set are shown in Table~\ref{sent2epa_examples},  along with their EPA predictions produced by \texttt{S2EPA}. We also include the closest word labels for each EPA from the ACT lexicon of behaviours.

\begin{table*}[!h]
\centering
\resizebox{0.8\textwidth}{!}{
\begin{tabular}{|l|l|l|}
\hline
{\bf Sentence} & {\bf EPA} & {\bf Closest ACT Labels}\\\hline\hline
i think i am in love & [1.60, 0.95, 0.55] & caution, collaborate with \\
\hline
 i hate you & [-1.63, 0.85, 0.49] & malign, injure \\
\hline
i have no fear of failure & [0.64, 1.27, 0.80] & train, confront \\
\hline
what the hell are you doing? & [-1.64, 0.41, 1.39] & badger, club \\
\hline
he's determined, unstoppable & [0.66, 1.87, 1.45] & apprehend, challenge\\
\hline
what do i do for fun? & [-0.35, -0.21, -0.04] & poke, gawk at\\
\hline
will you have dinner with me? & [0.91, 0.45, 0.79] & concur with, jest with \\
\hline
please don't talk with food in your mouth & [-0.82, 0.10, -0.64] & defer to, monitor\\ 
\hline
i insist on being told exactly what you have in mind & [0.06, 0.03, 0.13] & joggle, beckon to\\ 
\hline
you go ahead and relax, i'll cook & [0.95, 0.32, 0.47] & pay for, concur with\\
\hline
 i've been thinking about you & [1.59, 1.12, 0.66] & caution, collaborate with\\
 \hline
 you are despicable & [-1.74, 0.86, 0.94] & kick, club \\
 \hline
 i quit. & [-0.1, 0.89, 0.17] & search, smirk at\\
 \hline
 how about a drink? & [0.60, 0.42, 1.06] & query, jest with \\
 \hline
there is nothing for me here anymore & [-0.56, 0.30, 0.14] & flee, sound out \\
\hline
\end{tabular}
}
\caption{Examples of EPA vectors (and their closest word labels in ACT) produced for input sentences by \texttt{S2EPA}.}
\label{sent2epa_examples}
\end{table*}

We note that the model's EPA predictions are generally appropriate, and in many cases they are in alignment with the ACT behaviour labels. For instance, \textit{`i think i am in love'} is fairly positive due to the presence of the word \textit{love}; it is moderately potent and slightly active because of the phrase \textit{i think}. The closest labels in the ACT lexicon are \textit{caution} and \textit{collaborate with}. Among these, \textit{caution} seems to describe the input well.  A similar phenomenon is seen for the input \textit{`i hate you'}, whose EPA prediction closely matches the ACT labels \textit{malign, injure}. 
An interesting case is \textit{`i have no fear of failure'}: it has two negative and strong words \textit{fear} and \textit{failure}. Yet, the model correctly predicts that the overall sentiment of the response is positive and powerful, and is described well by the label \textit{confront}. 

We also see some negative examples. The E value of \textit{`i quit'} is $-0.1$, but it should be much more negative. The closest ACT labels \textit{search} and \textit{smirk at} don't make sense either. 
Similarly, the input \textit{`i've been thinking about you'} is composed of fairly neutral individual words; however the model correctly predicts that overall the sentence is positive, moderately potent and slightly active. 
On the other hand, its ACT labels \textit{caution} and \textit{collaborate with} don't seem altogether appropriate. 

To quantify the success rate of alignment between sentences and \texttt{S2EPA}'s predictions, we take a test set of 100 sentences (from the Cornell corpus, as before) and ask 3 human judges to classify the EPA predictions as `good' (1) or `bad' (0). Averaging these ratings yields a score of 0.739, indicating that 73.9\% of the EPA predictions (on the test set) are in alignment with the input sentence affect. Thus, overall the \texttt{S2EPA} function works reasonably well.

\subsubsection{Experiment \# 2}

Next, we analyse the affective quality of responses produced by the two variants of the \texttt{EPA2S} model: Seq2Seq and CVAE. 

First, we present the outputs of the traditional Seq2Seq model (without the $\bm \alpha$ vectors as input). 
Table~\ref{seq2seq_naive} shows its predictions on a small test set of inputs $\bm C$ chosen from the Cornell test set. We observe that the responses are grammatically well-formed but generally short and vague; this is a known shortcoming of naive Seq2Seq models.

\begin{table*}[!h]
\centering
\resizebox{0.55\textwidth}{!}{
\begin{tabular}{|l|l|}
\hline
{$\bm C$} & {\bf Traditional Seq2Seq (baseline)} \\\hline\hline
i hate you & that is okay \\ 
\hline
i think i'm in love & who is it  \\ 
\hline
i have no fear of failure & i have it in me \\ 
\hline
how about a drink? & sure why not \\ 
\hline
i've been thinking about you &  that makes it okay\\ 
\hline
there is nothing for me here anymore & so what do you want \\ 
\hline
you are despicable & so are you \\ 
\hline
what the hell are you doing? & i will get it for you \\ 
\hline
i quit & i don't think so \\ 
\hline
please don't talk with food in your mouth & sure if you want to \\ 
\hline
i insist on being told exactly what you have in mind & i have it all \\ 
\hline
\end{tabular}
}
\caption{The outputs of traditional Seq2Seq with attention, without $\bm \alpha$ labels.}
\label{seq2seq_naive}
\end{table*}

Next, we add an EPA vector $\bm \alpha$ to each training sample; in this experiment the $\bm \alpha$ is chosen manually from the ACT lexicon. 
Table~\ref{epa2sent_examples} shows the input prompts, target EPA vectors and the corresponding text generated by \texttt{EPA2S-Seq2Seq} and \texttt{EPA2S-CVAE}. 
\begin{table*}[!h]
\centering
\resizebox{0.9\textwidth}{!}{
\begin{tabular}{|l|l|l|l|l|}
\hline
\bf{Line \#} & {$\bm C$} & \textbf{Target} {$\bm \alpha$ \textbf{(Manually Chosen)}} & {\bf \texttt{EPA2S-Seq2Seq}} & {\bf \texttt{EPA2S-CVAE}} \\\hline\hline
1 &  & [1.71,1.39,-0.90] (calm) & you know me & what do you want \\ 
\cline{1-1}
\cline{3-5}
 2 & i hate you & [-0.50,0.72,0.81] (criticize) & okay & man can you scream \\
 \cline{1-1}
\cline{3-5}
 3 & & [-0.83,-0.93,0.44] (hide from) & you write a proper part for me & i feel so tired \\
\hline
4 &  & [0.98,0.38,0.02] (agree with) & who is it & i don't really know you \\ 
\cline{1-1}
\cline{3-5}
5 & i think i'm in & [-1.39,-0.47,2.15] (laugh at) & wait up please & yeah but don't make any noise\\
\cline{1-1}
\cline{3-5}
6 & love & [-1.53,-0.20,-0.19] (ignore) & i don't think so & we should find a leader to fight\\
\hline
7 &  & [2.14,1.21,-0.17] (appreciate) & yes & i believe it when you say \\ 
\cline{1-1}
\cline{3-5}
8 &  i have no fear & [-1.61,0.66,1.25] (antagonize) & i don't know & i need to leave early tomorrow\\
\cline{1-1}
\cline{3-5}
9 & of failure & [1.90,0.82,-0.11] (smile at) & what do you say & i know you, $<$unk$>$\\
\hline
10 &  & [0.98,0.38,0.02] (agree with) & sure that's nice & let me see what i can do about you \\ 
\cline{1-1}
\cline{3-5}
11 & how about a & [-1.05,-0.69,0.33] (avoid) & i'm sorry i can't & there is something on the clouds\\
\cline{1-1}
\cline{3-5}
12 & drink?  & [1.18,1.47,0.20] (charm) & how long have you been awake & i'm going with you baby \\
\hline
13 & there is nothing & [2.12,1.12,-0.81] (comfort) & yeah you know me & it is better this way \\ 
\cline{1-1}
\cline{3-5}
14 & for me here  & [1.64,1.17,0.47] (encourage) & no it is & it's not too late to try \\
\cline{1-1}
\cline{3-5}
15 & anymore &  [1.27,1.14,1.44] (entertain) & not now & you need to calm down \\
\hline
16 & & [-1.53,-0.20,-0.19] (ignore) & that is okay & man can you scream \\ 
\cline{1-1}
\cline{3-5}
17 & you are  & [-0.83,-0.93,0.44] (hide from) & i will not go to him  & you can show the way  \\
\cline{1-1}
\cline{3-5}
18 & despicable &  [1.71,1.39,-0.90] (calm) & i am your friend & these are great times we live in \\
\hline
19 &  & [2.12, 1.12, -0.81] (comfort) & i can go right now & i am singing for her \\ 
\cline{1-1}
\cline{3-5}
20 & what the hell  & [-1.05,-0.69,0.33] (avoid) & i can ask you & what do you want \\
\cline{1-1}
\cline{3-5}
21 & are you doing? &  [1.27,1.14,1.44] (entertain) & what do you think? & i won't mind a shower \\
\hline
22 &  & [2.12,1.12,-0.81] (comfort) & i don't think so  & you can do better than me \\ 
\cline{1-1}
\cline{3-5}
23 & i quit.  & [1.64,1.17,0.47] (encourage) & you know me & it's not too late to try \\
\cline{1-1}
\cline{3-5}
24 &  &  [-0.50,0.72,0.81] (criticize) & not now & who can help you \\
\hline
25 & please don't & [-1.39,-0.47,2.15] (laugh at) & i can do it & what do you want \\ 
\cline{1-1}
\cline{3-5}
26 & talk with food  & [-1.53,-0.20,-0.19] (ignore) & okay so & this is how it is \\
\cline{1-1}
\cline{3-5}
27 & in your mouth &  [1.18,1.47,0.20] (charm) & okay i will & that's okay for you \\
\hline
28 & i insist on being & [2.12, 1.12, -0.81] (comfort) & i can help you with that & i won't mind some baby \\ 
\cline{1-1}
\cline{3-5}
29 & told exactly what & [1.64, 1.17, 0.47] (encourage) & i am your friend & you can take it off  \\
\cline{1-1}
\cline{3-5}
30 & you have in mind &  [1.27, 1.14, 1.44] (entertain) & excited for you & please take it back \\
\hline
\end{tabular}
}
\caption{Example outputs generated by \texttt{EPA2S} for a given input sentence and EPA vector.}
\label{epa2sent_examples}
\end{table*}

Similar to the Seq2Seq baseline, we see short and non-committal responses by \texttt{EPA2S-Seq2Seq}. 
As far as their quality and relevance is concerned, we see some positive examples (Lines 1, 4, 6, 7, 10, 11, 14, 17, 18, 19, 23, 26, 28, 30) where the output sentences are well-aligned with the inputs $\bm C$ and $\bm \alpha$; the rest of the examples show output that is syntactically coherent but does not align well with either $\bm C$ or $\bm \alpha$ or both. For instance, in Line 2, \textit{`okay'} is a valid response to \textit{`i hate you'}, but it does not correspond to criticizing. Similarly, in Line 5, the response \textit{`wait up please'} is not relevant to the input \textit{`i think i'm in love'} or the target affect of \textit{laugh at}. Overall, the results are pretty evenly divided between positive and negative examples.

We see similar results for \texttt{EPA2S-CVAE}. There are some positive examples (Lines 2, 3, 7, 12, 13, 14, 16, 18--23, 26). On the other hand we see several outputs that are contextually relevant but affectively misaligned (Lines 1, 4, 5, 15, 24, 30). The responses are generally longer and less vague than baseline Seq2Seq and \texttt{EPA2S-Seq2Seq}. 

To quantify the performance of the two \texttt{EPA2S} variants, we set up an experiment as follows. Given a test set of 100 sentences and the target $\bm \alpha$ vector, we ask 3 human judges to specify whether the predicted response aligns with $\bm C$, $\bm \alpha$, both or none. The results are presented in Table~\ref{quant_epa2s}.
Overall, the results are evenly distributed  across the four classes. Strictly speaking, the success rate (alignment with both $\bm C$ and $\bm \alpha$) is 23.1\% and 27.6\% respectively for \texttt{EPA2S-Seq2Seq} and \texttt{EPA2S-CVAE}.

\begin{table*}[!h]
\centering
\resizebox{0.63\textwidth}{!}{
\begin{tabular}{|l|l|l|}
\hline
\bf{ } & {\bf \texttt{EPA2S-Seq2Seq}} & {\bf \texttt{EPA2S-CVAE}} \\\hline\hline
 \% Alignment with $\bm C$ and $\bm \alpha$ & 23.1 & 27.6 \\ 
\hline
 \% Alignment with $\bm C$ only & 25.5 & 22.0 \\ 
\hline
 \% Alignment with $\bm \alpha$ only & 22.6 & 20.7 \\ 
\hline
 \% Alignment with neither $\bm C$ nor $\bm \alpha$ & 28.8 & 29.7 \\ 
\hline
\end{tabular}
}
\caption{Evaluating the two \texttt{EPA2S} variants.}
\label{quant_epa2s}
\end{table*}

\begin{table*}[!t]
	\centering
	\resizebox{0.75\textwidth}{!}{
		\begin{tabular}{|c||c|c|c|} 
			\hline
			\multirow{2}{*}{\textbf{Model}} & {\textbf{Syntactic}} & \textbf{Natural-} & {\textbf{Emotional}}\\
            & \textbf{Coherence} & \textbf{ness} & \textbf{Approp.}\\
			\hline\hline
			\small{Traditional Seq2Seq (baseline)} & \hspace{-10pt} 1.48 & \hspace{-10pt} 0.69 & \hspace{-10pt} 0.41 \\
			\hline
			\small{ACT with \texttt{S2EPA} \& \texttt{EPA2S-Seq2Seq}} (\textit{friend-friend}) & 1.59 $\downarrow$ & 0.73 $\downarrow$ & 0.39 $\downarrow$ \\
			\hline
			\small{ACT with \texttt{S2EPA} \& \texttt{EPA2S-CVAE} (\textit{friend-friend})} & 1.57 $\downarrow$ & 0.68 $\downarrow$ & 0.47 $\downarrow$ \\
			\hline
			\small{ACT with \texttt{S2EPA} \& \texttt{EPA2S-Seq2Seq}} (\textit{enemy-enemy}) & 1.54 $\downarrow$ & 0.82 $\uparrow$ & 0.49 $\downarrow$ \\
			\hline
			\small{ACT with \texttt{S2EPA} \& \texttt{EPA2S-CVAE} (\textit{enemy-enemy})} & 1.55 $\downarrow$ & 0.73 $\downarrow$ & 0.59 $\uparrow$ \\
			\hline
		\end{tabular}
	}
		\caption{Comparing the different ACT conversation models. Arrows provide statistical significance of results. The up arrows indicate that the model's score is significantly better than the baseline ($p=0.05$). The down arrows indicate that the model's score is not significantly better than the baseline ($p=0.05$).}
		\label{tab:act_models}
\end{table*}

\begin{table*}[!h]
\centering
\resizebox{\textwidth}{!}{
\begin{tabular}{|c|l|l|c|l|l|}
\hline
{\bf Line} & {$\bm C$} & {\bf Target} $\bm \alpha$ {\bf (ACT) \& Closest ACT Labels} & {\bf Defl.} & {\bf \texttt{EPA2S-Seq2Seq}} & {\bf \texttt{EPA2S-CVAE}} \\\hline\hline
1 & i hate you & [2.52, 2.52, -0.41] (care for, caress) & 17.09 & that's not the point & you must be tired now \\ 
\hline
2 & i think i'm in love & [3.13, 1.70, 1.39] (thank, kiss) & 1.84 & i'm glad you like it & i wouldn't do you if i were you \\ 
\hline
3 & i have no fear of failure & [3.72, 1.90, 1.3] (thank, propose marriage to) & 4.36 & well that's me & i will ride with you love \\ 
\hline
4 & how about a drink? & [3.37, 1.68, 0.92] (reward, thank) & 4.06 & sure that's nice & i have money \\ 
\hline
5 & i've been thinking about you & [3.12, 1.96, 1.31] (thank, kiss) & 1.87 & okay & i like you \\ 
\hline
6 & there is nothing for me here anymore & [3.55, 1.99, 0.45] (embrace, propose marriage to) & 9.05 & i don't think so & it is better this way \\ 
\hline
\end{tabular}
}
\caption{The full ACT conversational model with ACT identities \textit{friend-friend}.}
\label{full_model_friend}
\end{table*}


\begin{table*}[!h]
\centering
\resizebox{0.99\textwidth}{!}{
\begin{tabular}{|c|l|l|c|l|l|}
\hline
\textbf{Line} & {$\bm C$} & {\bf Target} $\bm \alpha$ {\bf (ACT) and Closest ACT Labels} & \bf{Defl.} & {\bf \texttt{EPA2S-Seq2Seq}} & {\bf \texttt{EPA2S-CVAE}} \\\hline\hline
1 & i hate you & [-0.27, 0.35, 0.77] (bellow at) & 2.21 & i am not your friend & man can you scream \\ 
\hline
2 & you are despicable & [-0.18, 0.55, 0.58] (disagree with) & 4.32 & i don't care for you & you can calm down \\ 
\hline
3 & what the hell are you doing & [-0.29, 0.35, 0.74] (bellow at) & 3.56 & i can ask you it & i need to leave \\ 
\hline
4 & i quit. & [-0.17, 0.39, 0.75] (giggle at) & 5.29 & well that's me  & it is too late  \\ 
\hline
5 & please don't talk with food in your mouth & [-0.09, 0.48, 0.64] (disagree with) & 6.30 & not now & go away dog \\ 
\hline
6 & i insist on being told exactly what you have in mind & [-0.17, 0.33, 1.12] (be sarcastic toward) & 4.01 & yeah you know me & i am singing for you \\ 
\hline
\end{tabular}
}
\caption{The full ACT conversational model with ACT identities \textit{enemy-enemy}.}
\label{full_model_enemy}
\end{table*}



\subsubsection{Experiment \# 3}
We now test the full model (the dialogue pipeline shown in Figure~\ref{bactgen_pipeline}), where the two functions \texttt{S2EPA} and \texttt{EPA2S} are integrated with ACT. That is to say, the target EPA vectors $\bm \alpha$ are produced by ACT. We use two ACT settings for identities: \textit{friend-friend} and \textit{enemy-enemy}.

First, we quantitatively compare the quality of four variants of the ACT model with the baseline Seq2Seq model in Table~\ref{tab:act_models}. We ask three human judges to rate the responses of each model on 100 test prompts, given the affective identities of the two participants. The three evaluation axes are \textit{syntactic coherence} (Does the response
make grammatical sense?), \textit{naturalness} (Could the response have been plausibly produced
by a human?), and \textit{emotional appropriateness} (Is the response emotionally suitable for the
prompt?)~\cite{asghar2018affective}.  For each of these axes, the judges are asked to assign each response an integer score of
0 (bad), 1 (satisfactory), or 2 (good). The scores are then averaged for each axis. We also compute the statistical significance of the results using one-tailed Wilcoxon’s Signed Rank
Test~\cite{wilcoxon1945individual} with significance level set to 0.05. This is indicated through arrows in Table~\ref{tab:act_models}: a down-arrow indicates that the model performed equally well as the baseline, and an up-arrow indicates that the model performed significantly better than the
baseline. We see that all four models perform on par with the baseline, as far as syntactic coherence is concerned. \texttt{EPA2S-Seq2Seq}'s naturalness for the \textit{enemy-enemy} setting is significantly better than others, whereas \texttt{EPA2S-CVAE}'s emotional appropriateness is the highest for the \textit{enemy-enemy} setting (as indicated by arrows).

Next, we examine the results qualitatively.
We first analyse the setting where the ACT identity of both interactants is \textit{friend}. The results are shown in Table~\ref{full_model_friend}. We see that ACT produces target actions that are very friendly and nice (e.g. \textit{care for, thank, kiss, embrace}). This is consistent with the responder's identity of \textit{friend}. 
Both Seq2Seq and CVAE produce responses that are generally well-formed and relevant to the input prompt $\bm C$, but they sometimes seem to ignore $\bm \alpha$. Though the affective interpretation of the responses is very subjective, we observe that Seq2Seq produces emotionally aligned responses in Lines 2 and 4, while CVAE produces affectively appropriate results on Lines 1, 5 and 6. We also include the ACT deflection values in the table for the sake of completeness.

In the second setting, we set both the ACT identities to \textit{enemy}.
The results are presented in Table~\ref{full_model_enemy}. We observe that the actions predicted by ACT are not friendly anymore (\textit{giggle at, disagree with, bellow at, be sarcastic toward}); these behaviours are consistent with the responder's identity of \textit{enemy}. Here we see that the responses often align well with $\bm \alpha$ in many cases. The positive examples for Seq2Seq are Lines 1,2 5 and 6; those for CVAE are Lines 1, 2, 3 and 5, 6.

Overall, it can be concluded that the performance of ACT response generation is generally better than baseline models, as far as contextual relevance and emotional appropriateness are concerned. We do see some negative examples too: they can be attributed to the underwhelming performance of the \texttt{EPA2S} models for certain ACT identity settings.

\section{Discussion}
\label{bactgen_limitations}
Based on the three experiments presented in the previous section, our main takeaway is that \texttt{S2EPA} often performs reasonably well, whereas \texttt{EPA2S} may be more susceptible to the choice of ACT identities. In this work we chose two simple settings \textit{friend-friend} and \textit{enemy-enemy} as canonical examples. However, in the real world, identities are much more nuanced; this is also true for our training set of movie transcripts. In fact, a big chunk of the training set may not align well with either of the two settings. As an example, for the \textit{friend-friend} setting, we should only train using movie conversations that happen between two friendly identities (e.g. mother and child, two colleagues, two friendly spouses).  Thus, ACT identities need to be chosen more carefully and, once they are fixed, the appropriate training examples from the data should be used for training.

We highlight some other shortcomings of our models that contribute to partially negative results.
\begin{itemize}
    \item The sentence-level EPA vectors predicted by the \texttt{S2EPA} model may not be precisely accurate, as expected by ACT. Thus, even small discrepancies on the EPA scale can be detrimental to the CVAE or Seq2Seq learning. One way to alleviate this problem is to carry out a large-scale user study and construct a lexicon of sentence-EPA pairs, much like the word-level ACT lexicon. However, such a lexicon would have to be gathered for each identity pair independently. 
    \item For the \texttt{EPA2S} models, a dataset of $\sim$220k $(\bm C, \bm \alpha, \bm X)$ triples may be small enough to cause over-fitting. Ideally, for each $\bm C$, the data should contain examples with different $\bm \alpha$ vectors; this would allow the model to understand how the response affect should vary for a fixed $\bm C$. In turn, this would enable the model to control and capture global affective features more effectively. Constructing such a dataset may be time-consuming and expensive. Another possible approach may be to disentangle the multidimensional representations of affect and content, following \cite{hu2017toward,john2019disentangled}. In this case, generating the sentences may be less noisy and the dataset for training the model would not require as many examples.
    \item The process of converting EPA values to appropriate conversational responses is a hard problem in general, even for humans. For example, given $\bm C =$ \textit{`i failed my exam'} and $\bm \alpha=[1.97, 1.71, 1.51]$ (without a word label), it is not obvious how to come up with an appropriately worded, grammatically correct response that precisely conveys the right amount of evaluation, potency and activity. Furthermore, each EPA may correspond to many valid sentences, and each sentence may have many valid EPA ratings, due to the subjectivity of the task. 
\end{itemize}

\section{Conclusion}
We propose a neural conversational system that uses Affect Control Theory (ACT) to guide the generation of affective dialogue responses. In particular, we develop models that convert dialogue actions (i.e. sentences) to ACT actions (EPA vectors) and vice versa. We also discuss their relative strengths and weaknesses. The experiments generally show  positive results, and highlight some key limitations of the proposed models. We provide ideas about how the limitations can be addressed in the future.

\section{Acknowledgements}
We thank Olga Vechtomova and Ilya Shapiro for their feedback and insightful discussions.

Part of this work is presented in Chapter 4 of the first author's Ph.D. thesis~\cite{asghar2019emotion}.

\bibliographystyle{aaai}
\bibliography{aaai}

\end{document}

%% file: bayesactdefs.tex
\newcommand{\iagent}{{\em i-agent}\xspace}
\newcommand{\sagent}{{\em s-agent}\xspace}
\newcommand{\iagents}{{\em i-agents}\xspace}
\newcommand{\sagents}{{\em s-agents}\xspace}
\newcommand{\Bayesocial}{{\em Bayes.social}\xspace}
\newcommand{\chat}{{interaction}\xspace}
\newcommand{\chats}{{interactions}\xspace}
\newcommand{\Chat}{{Interaction}\xspace}
\newcommand{\Chats}{{Interactions}\xspace}
\newcommand{\msg}{{message}\xspace}
\newcommand{\msgs}{{messages}\xspace}
\newcommand{\Msg}{{Message}\xspace}
\newcommand{\Msgs}{{Messages}\xspace}
\newcommand{\robots}{robots\xspace}
\newcommand{\robot}{robot\xspace}
\newcommand{\member}{member\xspace}
\newcommand{\Member}{Member\xspace}
\newcommand{\members}{members\xspace}
\newcommand{\Members}{Members\xspace}
\newcommand{\amember}{a-member\xspace}
\newcommand{\Amember}{A-member\xspace}
\newcommand{\amembers}{a-members\xspace}
\newcommand{\Amembers}{A-members\xspace}
\newcommand{\connection}{connection\xspace}
\newcommand{\Connection}{Connection\xspace}
\newcommand{\connections}{connections\xspace}
\newcommand{\Connections}{Connections\xspace}
\newcommand{\friends}{friends\xspace}
\newcommand{\friendps}{friend's\xspace}
\newcommand{\Fs}{\mathrm{S_f}}
\newcommand{\fs}{\mathrm{s_f}}
\newcommand{\Ss}{\mathrm{S_s}}
\newcommand{\Ia}{\mathrm{I_a}}
\newcommand{\ia}{\mathrm{i_a}}
\renewcommand{\ia}{\mathrm{i}}

\newcommand{\Fsb}{\mathrm{\mathbf{S_f}}}
\newcommand{\fsb}{\mathrm{\mathbf{s_f}}}
\newcommand{\Ssb}{\mathrm{\mathbf{S_s}}}
\newcommand{\ssb}{\mathrm{\mathbf{s_s}}}
\newcommand{\Msb}{\mathrm{\mathbf{S_\circ}}}
\newcommand{\msb}{\mathrm{\mathbf{s_\circ}}}
\newcommand{\Iab}{\mathrm{\mathbf{I_a}}}
\newcommand{\iab}{\mathrm{\mathbf{i_a}}}
\newcommand{\Ib}{\mathrm{\mathbf{I}}}
\newcommand{\ib}{\mathrm{\mathbf{i}}}

\newcommand{\overbar}[1]{\overline{#1}}

\newcommand{\pastt}{\kappa}
\newcommand{\futt}{\tau}
\newcommand{\Tr}{\mathrm{T}}
\newcommand{\X}{\mathrm{X}}
\newcommand{\Fu}{\mathrm{F}}
\newcommand{\fu}{\mathrm{f}}
\newcommand{\tr}{\tau}
\newcommand{\Trb}{\mathrm{\mathbf{T}}}
\newcommand{\Xb}{\mathrm{\mathbf{X}}}
\newcommand{\Xbp}{\mathrm{\mathbf{X}'}}
\newcommand{\Xbb}{\mathrm{\mathbf{X_b}}}
\newcommand{\Xab}{\mathrm{\mathbf{X_a}}}
\newcommand{\Xcb}{\mathrm{\mathbf{X_c}}}
\newcommand{\Xbab}{\mathrm{\mathbf{X_{b_a}}}}
\newcommand{\Xbcb}{\mathrm{\mathbf{X_{b_c}}}}
\newcommand{\Lb}{\mathrm{\mathbf{L}}}
\newcommand{\Yb}{\mathrm{\mathbf{Y}}}
\newcommand{\Fub}{\mathrm{\mathbf{F}}}
\newcommand{\Gub}{\mathrm{\mathbf{\hat{G}}}}
\newcommand{\Gube}{\mathrm{\mathbf{\hat{G}_{\epsilon}}}}
\newcommand{\Aab}{\mathrm{\mathbf{B_a}}}
\newcommand{\Emb}{\mathrm{\mathbf{B_e}}}
\newcommand{\Bb}{\mathrm{\mathbf{B}}}
\newcommand{\bb}{\mathrm{\mathbf{b}}}
\newcommand{\Aa}{\mathrm{B_a}}
\newcommand{\ab}{\mathrm{\mathbf{a}}}
\newcommand{\aeb}{\mathrm{\mathbf{a_{\epsilon}}}}
\newcommand{\aab}{\mathrm{\mathbf{b_a}}}
\newcommand{\emb}{\mathrm{\mathbf{b_e}}}
\newcommand{\ba}{\mathrm{b_a}}
\newcommand{\state}{\bm{s}}
\newcommand{\Thfb}{\bm{\Theta_f}}
\newcommand{\Thb}{\bm{\Theta_b}}
\newcommand{\thb}{\bm{\theta_b}}
\newcommand{\thfb}{\bm{\theta_f}}
\newcommand{\Om}{\bm{\Omega}}
\newcommand{\om}{\bm{\omega}}
\newcommand{\Omb}{\bm{\Omega_f}}
\newcommand{\omb}{\bm{\omega_f}}
\newcommand{\Ombe}{\bm{\Omega_{\epsilon}}}
\newcommand{\ombe}{\bm{\omega_{\epsilon}}}
\newcommand{\omeab}{\bm{\omega_{\epsilon_a}}}
\newcommand{\omecb}{\bm{\omega_{\epsilon_c}}}
\newcommand{\Omfb}{\bm{\Omega_f}}
\newcommand{\omfb}{\bm{\omega_f}}
\newcommand{\Omeb}{\bm{\Omega_e}}
\newcommand{\omeb}{\bm{\omega_e}}
\newcommand{\Omxb}{\bm{\Omega_x}}
\newcommand{\omxb}{\bm{\omega_x}}
\newcommand{\Omsb}{\bm{\Omega_s}}
\newcommand{\omsb}{\bm{\omega_s}}
\newcommand{\ombcb}{\bm{\omega_{b_c}}}
\newcommand{\mub}{\bm{\mu}}
\newcommand{\mubn}{\mub_n}
\newcommand{\Sigb}{\bm{\Sigma}}
\newcommand{\Sigbb}{\Sigb_b}
\newcommand{\Sigbg}{\Sigb_g}
\newcommand{\Sigbn}{\Sigb_n}
\newcommand{\Sigbf}{\Sigb_f}
\newcommand{\alphab}{\mathrm{\mathbf{\alpha}}}
\newcommand{\betab}{\mathrm{\mathbf{\beta}}}
\newcommand{\Omx}{\Omega_x}
\newcommand{\omx}{\omega_x}
\newcommand{\Omf}{\Omega_f}
\newcommand{\ome}{\omega_e}
\newcommand{\Ome}{\Omega_e}
\newcommand{\omf}{\omega_f}
\newcommand{\Tha}{\Theta_b}
\newcommand{\tha}{\theta_b}
\newcommand{\Thf}{\Theta_b}
\newcommand{\thf}{\theta_b}
\newcommand{\fub}{\mathrm{\mathbf{f}}}
\newcommand{\gub}{\mathrm{\mathbf{g}}}
\newcommand{\emotb}{\mathrm{\mathbf{\epsilon}}}
\newcommand{\emotab}{\mathrm{\mathbf{\epsilon_a}}}
\newcommand{\emotcb}{\mathrm{\mathbf{\epsilon_c}}}
\newcommand{\Emotb}{\mathrm{\mathbf{E}}}
\newcommand{\emot}{\mathrm{\epsilon}}
\newcommand{\Emot}{\mathrm{E}}
\newcommand{\Epa}{\bf{\mathcal{Y}}}
\newcommand{\ActSpace}{\bm{\mathcal{A}}}
\newcommand{\SenSpace}{\bm{\mathcal{G}}}
\newcommand{\ConSpace}{\bm{\mathcal{Y}}}
\newcommand{\DenSpace}{\bm{\mathcal{X}}}
\newcommand{\StateSpace}{\bm{\mathcal{S}}}

\newcommand{\evidence}{\mathbf{e}}
\newcommand{\gc}{\gamma^{\circ}}
\newcommand{\gp}{\gamma_p}

\newcommand{\threeids}{\mathbb{I}_{1,3}}
\newcommand{\diag}{\mathop{\mathrm{diag}}}

\newcommand{\bprod}[2]{\langle#1,#2\rangle}
\newcommand{\fubt}{\langle\fub,\aab\rangle}
\newcommand{\trb}{\bm{\tau}}
\newcommand{\xb}{\mathrm{\mathbf{x}}}
\newcommand{\xab}{\mathrm{\mathbf{x_a}}}
\newcommand{\xbb}{\mathrm{\mathbf{x_b}}}
\newcommand{\xbab}{\mathrm{\mathbf{x_{b_a}}}}
\newcommand{\xbcb}{\mathrm{\mathbf{x_{b_c}}}}
\newcommand{\xcb}{\mathrm{\mathbf{x_c}}}
\newcommand{\xxxb}{\mathrm{\mathbf{x_x}}}
\newcommand{\xxxbp}{\mathrm{\mathbf{x_x}'}}
\newcommand{\Xxxb}{\mathrm{\mathbf{X_x}}}
\newcommand{\Xxxbp}{\mathrm{\mathbf{X_x}'}}
\newcommand{\xeb}{\mathrm{\mathbf{x_{\epsilon}}}}
\newcommand{\xeab}{\mathrm{\mathbf{x_{\epsilon_a}}}}
\newcommand{\xecb}{\mathrm{\mathbf{x_{\epsilon_c}}}}
\newcommand{\yb}{\mathrm{\mathbf{y}}}
\newcommand{\x}{\mathrm{x}}
\newcommand{\ACTf}{\mathscr{L}}
\newcommand{\Gop}{\mathscr{G}}
\newcommand{\Xop}{\mathscr{X}}
\newcommand{\ACTM}{\bm{M}}
\newcommand{\ACTt}{\bm{t}}
\newcommand{\ACTH}{\mathscr{H}}
\newcommand{\ACTK}{\mathscr{K}}
\newcommand{\ACTc}{\mathscr{C}}
\newcommand{\ACTg}{\mathscr{J}}
\newcommand{\APPx}{\mathscr{X}}
\newcommand{\APPy}{\mathscr{Y}}
\newcommand{\xxa}{\ACTH_a}
\newcommand{\xxb}{1-\ACTH_b}
\newcommand{\xxc}{\ACTH_c}
\newcommand{\xxbi}{(1-\ACTH_b)^{-1}}
\newcommand{\negspace}{\!\!\!\!\!\!\!\!\!\!\!\!\!\!\!\!\!\!}
\newcommand{\negspacemed}{\!\!\!\!\!\!\!\!\!\!\!\!\!\!\!\!}
\newcommand{\negspacesm}{\!\!\!\!\!\!\!\!\!}

\newcommand{\agentemb}[1]{{\em #1}}
\newcommand{\actor}{\agentemb{actor}\xspace}
\newcommand{\actors}{\agentemb{actors}\xspace}
\newcommand{\agent}{\agentemb{agent}\xspace}
\newcommand{\Agent}{\agentemb{Agent}\xspace}
\newcommand{\agents}{\agentemb{agents}\xspace}
\newcommand{\agentps}{\agentemb{agent'}s\xspace}
\newcommand{\agentsp}{\agentemb{agents'}\xspace}
\newcommand{\client}{\agentemb{client}\xspace}
\newcommand{\Client}{\agentemb{Client}\xspace}
\newcommand{\clients}{\agentemb{clients}\xspace}
\newcommand{\clientps}{\agentemb{client'}s\xspace}
\newcommand{\clientsp}{\agentemb{clients'}\xspace}

\newcommand{\bact}{{\em BayesAct}\xspace}
\newcommand{\occ}{{\em OCC}\xspace}
\newcommand{\bacts}{{\em BayesAct-S}\xspace}
\newcommand{\act}{{\em ACT}\xspace}
\newcommand{\acts}{{\em ACT-S}\xspace}

\newcommand{\nacb}{$\pi^{\dagger}$}

\newcommand{\interact}{{\em Interact}\xspace}

\newcommand{\onechat}[1]{{\em ``#1''}}

\newcommand{\dpagent}{{\em PD-Agent}\xspace}
\newcommand{\dpagents}{{\em PD-Agents}\xspace}

\newcommand{\pda}{{\em pdA}\xspace}
\newcommand{\pdc}{{\em pdC}\xspace}
\newcommand{\pdagent}{{\em pd-agent}\xspace}
\newcommand{\pdagents}{{\em pd-agents}\xspace}
\newcommand{\Pdagent}{{\em PD-agent}\xspace}
\newcommand{\Pdagents}{{\em PD-agents}\xspace}

\newcommand{\self}{{\em{self}}\xspace}
\newcommand{\elab}[1]{{\em{#1}}}
\newcommand{\epa}[3]{(EPA:$\{#1,#2,#3\}$)}
\newcommand{\selfepa}[3]{(EPA of self:$\{#1,#2,#3\}$)}
\newcommand{\epalike}[3]{{(EPA$\sim\{#1,#2,#3\}$)}\xspace}
\newcommand{\epaequals}[3]{EPA$=\!\!\!\{#1,#2,#3\}$\xspace}
\newcommand{\cstrat}[1]{{\bf (#1)}}
\newcommand{\epaid}[1]{{\em #1}}

\newcommand{\actlabel}[1]{}
\newcommand{\actclientsez}[1]{{\color{blue}{\bf\em #1}}}
\newcommand{\actagentsez}[1]{{\color{red}{\bf\em #1}}}

\newcommand{\oracle}{\pi_{heur}}

\newcommand{\homops}{{\em homo sociopsychologicus}\xspace}
\newcommand{\homopsc}{{\em homo sociopsychologicus populus}\xspace}
\newcommand{\homopsi}{{\em homo sociopsychologicus singula}\xspace}
\newcommand{\homoecon}{{\em homo economicus}\xspace}
\newcommand{\homos}{{\em homo sociologicus}\xspace}
\newcommand{\homoas}{{\em homo asinus}\xspace}

\newcommand\minus{%
  \setbox0=\hbox{-}%
  \vcenter{%
    \hrule width\wd0 height \the\fontdimen8\textfont3%
  }%
}

\newcommand{\mi}{\minus}

\makeatletter
\def\grd@save@target#1{%
  \def\grd@target{#1}}
\def\grd@save@start#1{%
  \def\grd@start{#1}}

\makeatother

%% file: formatting-instructions-latex-2020.bbl
\begin{thebibliography}{}

\bibitem[\protect\citeauthoryear{Alhothali and Hoey}{2017}]{alhothali2017semi}
Alhothali, A., and Hoey, J.
\newblock 2017.
\newblock Semi-supervised affective meaning lexicon expansion using semantic
  and distributed word representations.
\newblock {\em arXiv preprint arXiv:1703.09825}.

\bibitem[\protect\citeauthoryear{Asghar and Hoey}{2015}]{asghar2015intelligent}
Asghar, N., and Hoey, J.
\newblock 2015.
\newblock Intelligent affect: Rational decision making for socially aligned
  agents.
\newblock In {\em UAI},  12--16.

\bibitem[\protect\citeauthoryear{Asghar \bgroup et al\mbox.\egroup
  }{2018}]{asghar2018affective}
Asghar, N.; Poupart, P.; Hoey, J.; Jiang, X.; and Mou, L.
\newblock 2018.
\newblock Affective neural response generation.
\newblock In {\em European Conference on Information Retrieval},  154--166.
\newblock Springer.

\bibitem[\protect\citeauthoryear{Asghar}{2019}]{asghar2019emotion}
Asghar, N.
\newblock 2019.
\newblock Emotion-aware and human-like autonomous agents.
\newblock {\em Ph.D. Thesis, Cheriton School of Computer Science, University of
  Waterloo}.

\bibitem[\protect\citeauthoryear{Bowman \bgroup et al\mbox.\egroup
  }{2016}]{bowman2016generating}
Bowman, S.~R.; Vilnis, L.; Vinyals, O.; Dai, A.; Jozefowicz, R.; and Bengio, S.
\newblock 2016.
\newblock Generating sentences from a continuous space.
\newblock In {\em CoNLL},  10--21.

\bibitem[\protect\citeauthoryear{Callejas, Griol, and
  L{\'o}pez-C{\'o}zar}{2011}]{callejas2011predicting}
Callejas, Z.; Griol, D.; and L{\'o}pez-C{\'o}zar, R.
\newblock 2011.
\newblock Predicting user mental states in spoken dialogue systems.
\newblock {\em EURASIP J. Advances in Signal Processing} 2011(1):6.

\bibitem[\protect\citeauthoryear{Catania \bgroup et al\mbox.\egroup
  }{2019}]{catania2019emoty}
Catania, F.; Di~Nardo, N.; Garzotto, F.; and Occhiuto, D.
\newblock 2019.
\newblock Emoty: An emotionally sensitive conversational agent for people with
  neurodevelopmental disorders.
\newblock In {\em Proceedings of the 52nd Hawaii International Conference on
  System Sciences}.

\bibitem[\protect\citeauthoryear{Danescu-Niculescu-Mizil and
  Lee}{2011}]{cornellcorpus}
Danescu-Niculescu-Mizil, C., and Lee, L.
\newblock 2011.
\newblock Chameleons in imagined conversations: A new approach to understanding
  coordination of linguistic style in dialogs.
\newblock In {\em Workshop on Cognitive Modeling and Computational
  Linguistics}.
\newblock Association for Computational Linguistics.

\bibitem[\protect\citeauthoryear{Dryja{\'n}ski \bgroup et al\mbox.\egroup
  }{2018}]{dryjanski2018affective}
Dryja{\'n}ski, T.; Bujnowski, P.; Choi, H.; Podlaska, K.; Michalski, K.; Beksa,
  K.; and Kubik, P.
\newblock 2018.
\newblock Affective natural language generation by phrase insertion.
\newblock In {\em IEEE International Conference on Big Data},  4876--4882.

\bibitem[\protect\citeauthoryear{Ekman}{1992}]{ekman1992argument}
Ekman, P.
\newblock 1992.
\newblock An argument for basic emotions.
\newblock {\em Cognition \& emotion} 6(3-4):169--200.

\bibitem[\protect\citeauthoryear{Felbo \bgroup et al\mbox.\egroup
  }{2017}]{felbo2017using}
Felbo, B.; Mislove, A.; S{\o}gaard, A.; Rahwan, I.; and Lehmann, S.
\newblock 2017.
\newblock Using millions of emoji occurrences to learn any-domain
  representations for detecting sentiment, emotion and sarcasm.
\newblock In {\em EMNLP},  1615--1625.

\bibitem[\protect\citeauthoryear{Fung \bgroup et al\mbox.\egroup
  }{2018}]{fung2018empathetic}
Fung, P.; Bertero, D.; Xu, P.; Park, J.~H.; Wu, C.-S.; and Madotto, A.
\newblock 2018.
\newblock Empathetic dialog systems.
\newblock In {\em LREC}.

\bibitem[\protect\citeauthoryear{Ghandeharioun \bgroup et al\mbox.\egroup
  }{2018}]{ghandeharioun2018emma}
Ghandeharioun, A.; McDuff, D.; Czerwinski, M.; and Rowan, K.
\newblock 2018.
\newblock Emma: An emotionally intelligent personal assistant for improving
  wellbeing.
\newblock {\em arXiv preprint arXiv:1812.11423}.

\bibitem[\protect\citeauthoryear{Ghosh \bgroup et al\mbox.\egroup
  }{2017}]{ghosh2017affectlm}
Ghosh, S.; Chollet, M.; Laksana, E.; Morency, L.-P.; and Scherer, S.
\newblock 2017.
\newblock {Affect-LM: A} neural language model for customizable affective text
  generation.
\newblock In {\em ACL}.

\bibitem[\protect\citeauthoryear{Gordon \bgroup et al\mbox.\egroup
  }{2019}]{gordon2019primer}
Gordon, C.; Leuski, A.; Benn, G.; Klassen, E.; Fast, E.; Liewer, M.; Hartholt,
  A.; and Traum, D.
\newblock 2019.
\newblock Primer: An emotionally aware virtual agent.
\newblock In {\em Proceedings of the IUI Workshop on User-Aware Conversational
  Agents}.

\bibitem[\protect\citeauthoryear{Hasegawa \bgroup et al\mbox.\egroup
  }{2013}]{hasegawa2013predicting}
Hasegawa, T.; Kaji, N.; Yoshinaga, N.; and Toyoda, M.
\newblock 2013.
\newblock Predicting and eliciting addressee’s emotion in online dialogue.
\newblock In {\em ACL (Volume 1: Long Papers)},  964--972.

\bibitem[\protect\citeauthoryear{Heise}{1979}]{Heise1979}
Heise, D.~R.
\newblock 1979.
\newblock {\em Understanding Events: Affect and the Construction of Social
  Action}.
\newblock New York: Cambridge University Press.

\bibitem[\protect\citeauthoryear{Heise}{2007}]{Heise2007}
Heise, D.~R.
\newblock 2007.
\newblock {\em Expressive Order: Confirming Sentiments in Social Actions}.
\newblock Springer.

\bibitem[\protect\citeauthoryear{Heise}{2010}]{Heise2010}
Heise, D.~R.
\newblock 2010.
\newblock {\em Surveying Cultures: Discovering Shared Conceptions and
  Sentiments}.
\newblock Wiley.

\bibitem[\protect\citeauthoryear{Hu \bgroup et al\mbox.\egroup
  }{2017}]{hu2017toward}
Hu, Z.; Yang, Z.; Liang, X.; Salakhutdinov, R.; and Xing, E.~P.
\newblock 2017.
\newblock Toward controlled generation of text.
\newblock In {\em ICML},  1587--1596.

\bibitem[\protect\citeauthoryear{Huang \bgroup et al\mbox.\egroup
  }{2017}]{huang2017moodswipe}
Huang, C.-Y.; Labetoulle, T.; Huang, T.-H.; Chen, Y.-P.; Chen, H.-C.;
  Srivastava, V.; and Ku, L.-W.
\newblock 2017.
\newblock Moodswipe: A soft keyboard that suggests messagebased on
  user-specified emotions.
\newblock In {\em EMNLP: System Demonstrations},  73--78.

\bibitem[\protect\citeauthoryear{Jaques \bgroup et al\mbox.\egroup
  }{2017}]{jaques2017multimodal}
Jaques, N.; Taylor, S.; Sano, A.; and Picard, R.
\newblock 2017.
\newblock Multimodal autoencoder: A deep learning approach to filling in
  missing sensor data and enabling better mood prediction.
\newblock In {\em 2017 Seventh International Conference on Affective Computing
  and Intelligent Interaction (ACII)},  202--208.

\bibitem[\protect\citeauthoryear{John \bgroup et al\mbox.\egroup
  }{2019}]{john2019disentangled}
John, V.; Mou, L.; Bahuleyan, H.; and Vechtomova, O.
\newblock 2019.
\newblock Disentangled representation learning for non-parallel text style
  transfer.
\newblock In {\em ACL},  424--434.

\bibitem[\protect\citeauthoryear{Kingma and Ba}{2015}]{adam}
Kingma, D., and Ba, J.
\newblock 2015.
\newblock Adam: A method for stochastic optimization.
\newblock In {\em ICLR}.

\bibitem[\protect\citeauthoryear{Kingma and Welling}{2013}]{kingma2013auto}
Kingma, D.~P., and Welling, M.
\newblock 2013.
\newblock Auto-encoding variational bayes.
\newblock {\em arXiv preprint arXiv:1312.6114}.

\bibitem[\protect\citeauthoryear{Kong \bgroup et al\mbox.\egroup
  }{2019}]{kong2019adversarial}
Kong, X.; Li, B.; Neubig, G.; Hovy, E.; and Yang, Y.
\newblock 2019.
\newblock An adversarial approach to high-quality, sentiment-controlled neural
  dialogue generation.
\newblock {\em arXiv preprint arXiv:1901.07129}.

\bibitem[\protect\citeauthoryear{Kort, Reilly, and
  Picard}{2001}]{kort2001affective}
Kort, B.; Reilly, R.; and Picard, R.~W.
\newblock 2001.
\newblock An affective model of interplay between emotions and learning:
  Reengineering educational pedagogy-building a learning companion.
\newblock In {\em IEEE International Conference on Advanced Learning
  Technologies},  43--46.

\bibitem[\protect\citeauthoryear{Liu \bgroup et al\mbox.\egroup
  }{2016}]{liu2016hownotto}
Liu, C.-W.; Lowe, R.; Serban, I.; Noseworthy, M.; Charlin, L.; and Pineau, J.
\newblock 2016.
\newblock How not to evaluate your dialogue system: An empirical study of
  unsupervised evaluation metrics for dialogue response generation.
\newblock In {\em EMNLP},  2122--2132.

\bibitem[\protect\citeauthoryear{Lubis \bgroup et al\mbox.\egroup
  }{2018}]{lubis2018eliciting}
Lubis, N.; Sakti, S.; Yoshino, K.; and Nakamura, S.
\newblock 2018.
\newblock Eliciting positive emotion through affect-sensitive dialogue response
  generation: A neural network approach.
\newblock In {\em AAAI}.

\bibitem[\protect\citeauthoryear{Malhotra \bgroup et al\mbox.\egroup
  }{2015}]{malhotra2015exploratory}
Malhotra, A.; Yu, L.; Schr{\"o}der, T.; and Hoey, J.
\newblock 2015.
\newblock An exploratory study into the use of an emotionally aware cognitive
  assistant.
\newblock In {\em AAAI Workshop: Artificial Intelligence Applied to Assistive
  Technologies and Smart Environments}.

\bibitem[\protect\citeauthoryear{McKeown \bgroup et al\mbox.\egroup
  }{2012}]{mckeown2012semaine}
McKeown, G.; Valstar, M.; Cowie, R.; Pantic, M.; and Schroder, M.
\newblock 2012.
\newblock The semaine database: Annotated multimodal records of emotionally
  colored conversations between a person and a limited agent.
\newblock {\em IEEE Transactions on Affective Computing} 3(1):5--17.

\bibitem[\protect\citeauthoryear{Mou \bgroup et al\mbox.\egroup
  }{2016}]{mou2016sequence}
Mou, L.; Song, Y.; Yan, R.; Li, G.; Zhang, L.; and Jin, Z.
\newblock 2016.
\newblock Sequence to backward and forward sequences: A content-introducing
  approach to generative short-text conversation.
\newblock In {\em COLING},  3349--3358.

\bibitem[\protect\citeauthoryear{Osgood, May, and Miron}{1975}]{Osgood1975}
Osgood, C.~E.; May, W.~H.; and Miron, M.~S.
\newblock 1975.
\newblock {\em Cross-Cultural Universals of Affective Meaning}.
\newblock University of Illinois Press.

\bibitem[\protect\citeauthoryear{Park}{2018}]{park2018finding}
Park, J.~H.
\newblock 2018.
\newblock Finding good representations of emotions for text classification.
\newblock {\em arXiv preprint arXiv:1808.07235}.

\bibitem[\protect\citeauthoryear{Pennington, Socher, and
  Manning}{2014}]{pennington2014glove}
Pennington, J.; Socher, R.; and Manning, C.~D.
\newblock 2014.
\newblock {GloVe}: Global vectors for word representation.
\newblock In {\em EMNLP},  1532--1543.

\bibitem[\protect\citeauthoryear{Pittermann, Pittermann, and
  Minker}{2010}]{pittermann2010emotion}
Pittermann, J.; Pittermann, A.; and Minker, W.
\newblock 2010.
\newblock Emotion recognition and adaptation in spoken dialogue systems.
\newblock {\em Int. J. Speech Technology} 13(1):49--60.

\bibitem[\protect\citeauthoryear{Plutchik}{1980}]{plutchik1980general}
Plutchik, R.
\newblock 1980.
\newblock A general psychoevolutionary theory of emotion.
\newblock In {\em Theories of emotion}. Elsevier.
\newblock  3--33.

\bibitem[\protect\citeauthoryear{Prendinger and
  Ishizuka}{2005}]{prendinger2005empathic}
Prendinger, H., and Ishizuka, M.
\newblock 2005.
\newblock The empathic companion: A character-based interface that addresses
  users'affective states.
\newblock {\em Applied Artificial Intelligence} 19(3-4):267--285.

\bibitem[\protect\citeauthoryear{Rashkin \bgroup et al\mbox.\egroup
  }{2018}]{rashkin2018know}
Rashkin, H.; Smith, E.~M.; Li, M.; and Boureau, Y.-L.
\newblock 2018.
\newblock I know the feeling: Learning to converse with empathy.
\newblock {\em arXiv preprint arXiv:1811.00207}.

\bibitem[\protect\citeauthoryear{Russell and Mehrabian}{1977}]{Russell1977}
Russell, J.~A., and Mehrabian, A.
\newblock 1977.
\newblock Evidence for a three-factor theory of emotions.
\newblock {\em Journal of research in Personality} 11(3):273--294.

\bibitem[\protect\citeauthoryear{Russell}{2003}]{Russell2003}
Russell, J.~A.
\newblock 2003.
\newblock Core affect and the psychological construction of emotion.
\newblock {\em Psychological Review} 110(1):145--172.

\bibitem[\protect\citeauthoryear{Serban \bgroup et al\mbox.\egroup
  }{2017}]{serban2017hierarchical}
Serban, I.~V.; Sordoni, A.; Lowe, R.; Charlin, L.; Pineau, J.; Courville,
  A.~C.; and Bengio, Y.
\newblock 2017.
\newblock A hierarchical latent variable encoder-decoder model for generating
  dialogues.
\newblock In {\em AAAI},  3295--3301.

\bibitem[\protect\citeauthoryear{Shang, Lu, and Li}{2015}]{shang2015neural}
Shang, L.; Lu, Z.; and Li, H.
\newblock 2015.
\newblock Neural responding machine for short-text conversation.
\newblock In {\em ACL-IJCNLP},  1577--1586.

\bibitem[\protect\citeauthoryear{Shen \bgroup et al\mbox.\egroup
  }{2017}]{shen2017conditional}
Shen, X.; Su, H.; Li, Y.; Li, W.; Niu, S.; Zhao, Y.; Aizawa, A.; and Long, G.
\newblock 2017.
\newblock A conditional variational framework for dialog generation.
\newblock In {\em ACL (Volume 2: Short Papers)},  504--509.

\bibitem[\protect\citeauthoryear{Shi and Yu}{2018}]{shi2018sentiment}
Shi, W., and Yu, Z.
\newblock 2018.
\newblock Sentiment adaptive end-to-end dialog systems.
\newblock In {\em ACL (Volume 1: Long Papers)},  1509--1519.

\bibitem[\protect\citeauthoryear{Smith \bgroup et al\mbox.\egroup
  }{2006}]{Japan2002Data}
Smith, H.; Matsuno, T.; Ike, S.; and Umino, M.
\newblock 2006.
\newblock Mean affective ratings of 1,894 concepts by japanese undergraduates,
  1989-2002.
\newblock Distributed at Affect Control Theory Website, Program Interact
  http://www.indiana.edu/~socpsy/ACT/interact/JavaInteract.html.

\bibitem[\protect\citeauthoryear{Sohn, Lee, and Yan}{2015}]{sohn2015learning}
Sohn, K.; Lee, H.; and Yan, X.
\newblock 2015.
\newblock Learning structured output representation using deep conditional
  generative models.
\newblock In {\em NIPS},  3483--3491.

\bibitem[\protect\citeauthoryear{Sutskever, Vinyals, and
  Le}{2014}]{sutskever2014sequence}
Sutskever, I.; Vinyals, O.; and Le, Q.~V.
\newblock 2014.
\newblock Sequence to sequence learning with neural networks.
\newblock In {\em NIPS},  3104--3112.

\bibitem[\protect\citeauthoryear{Taylor \bgroup et al\mbox.\egroup
  }{2017}]{taylor2017personalized}
Taylor, S.~A.; Jaques, N.; Nosakhare, E.; Sano, A.; and Picard, R.
\newblock 2017.
\newblock Personalized multitask learning for predicting tomorrow's mood,
  stress, and health.
\newblock {\em IEEE Transactions on Affective Computing}.

\bibitem[\protect\citeauthoryear{Vadehra}{2018}]{vadehra2018creating}
Vadehra, A.
\newblock 2018.
\newblock Creating an emotion responsive dialogue system.
\newblock Master's thesis, University of Waterloo.

\bibitem[\protect\citeauthoryear{Vinyals and Le}{2015}]{vinyals2015neural}
Vinyals, O., and Le, Q.
\newblock 2015.
\newblock A neural conversational model.
\newblock {\em arXiv preprint arXiv:1506.05869}.

\bibitem[\protect\citeauthoryear{Wilcoxon}{1945}]{wilcoxon1945individual}
Wilcoxon, F.
\newblock 1945.
\newblock Individual comparisons by ranking methods.
\newblock {\em Biometrics Bulletin} 1(6):80--83.

\bibitem[\protect\citeauthoryear{Xie \bgroup et al\mbox.\egroup
  }{2016}]{xie2016neural}
Xie, R.; Liu, Z.; Yan, R.; and Sun, M.
\newblock 2016.
\newblock Neural emoji recommendation in dialogue systems.
\newblock {\em arXiv preprint arXiv:1612.04609}.

\bibitem[\protect\citeauthoryear{Zhang and Wang}{2018}]{zhang2018learning}
Zhang, R., and Wang, Z.
\newblock 2018.
\newblock Learning to converse emotionally like humans: A conditional
  variational approach.
\newblock In {\em CCF International Conference on Natural Language Processing
  and Chinese Computing},  98--109.

\bibitem[\protect\citeauthoryear{Zhou and Wang}{2018}]{zhou2018mojitalk}
Zhou, X., and Wang, W.~Y.
\newblock 2018.
\newblock Mojitalk: Generating emotional responses at scale.
\newblock In {\em ACL (Volume 1: Long Papers)},  1128--1137.

\bibitem[\protect\citeauthoryear{Zhou \bgroup et al\mbox.\egroup
  }{2017}]{zhou2017ecm}
Zhou, H.; Huang, M.; Zhang, T.; Zhu, X.; and Liu, B.
\newblock 2017.
\newblock Emotional chatting machine: Emotional conversation generation with
  internal and external memory.
\newblock {\em arXiv preprint arXiv:1704.01074}.

\bibitem[\protect\citeauthoryear{Zhou \bgroup et al\mbox.\egroup
  }{2018}]{zhou2018design}
Zhou, L.; Gao, J.; Li, D.; and Shum, H.-Y.
\newblock 2018.
\newblock The design and implementation of xiaoice, an empathetic social
  chatbot.
\newblock {\em arXiv preprint arXiv:1812.08989}.

\end{thebibliography}
